\documentclass[nofootinbib,twocolumn,pre,superscriptaddress]{revtex4-1}

\makeatletter
\newif\if@restonecol
\makeatother

\usepackage{newfloat,algcompatible}
\usepackage[size=small]{caption}
\usepackage{etoolbox}

\AtEndEnvironment{algorithm}{\noindent\hrulefill\par\nobreak\vskip-5pt}

\usepackage{newfloat}
\DeclareFloatingEnvironment[
    fileext=loa,
    listname=List of Algorithms,
    name=ALGORITHM,
    placement=tbhp,
]{algorithm}

\DeclareCaptionFormat{algorithms}{\vskip-15pt\hrulefill\par#1#2#3\vskip-6pt\hrulefill}
\algblock[Input]{Input}{EndInput}
\algblockdefx[Input]{Input}{EndInput}%
    [1]{\textbf{Input} #1}%
    {}

\usepackage{graphicx}
\usepackage{amssymb}
\usepackage{amsmath}
\usepackage{amsfonts}
\usepackage{balance}
\usepackage{subfig}
\usepackage{hyperref}
\usepackage{bm}

\newcommand*\dif{\,\mathrm{d}} 										

\begin{document}

\title{Active Discovery of Network Roles for Predicting the Classes of Network Nodes}
\author{Leto Peel}
\email{leto.peel@colorado.edu}
\affiliation{Department of Computer Science, University of Colorado, Boulder, CO 80309}

\begin{abstract}
Nodes in real world networks often have class labels, or underlying attributes, that are related to the way in which they connect to other nodes.  Sometimes this relationship is simple, for instance nodes of the same class are may be more likely to be connected.  In other cases, however, this is not true,  and the way that nodes link in a network exhibits a different, more complex relationship to their attributes.  
Here, we consider networks in which we know how the nodes are connected, but we do not know the class labels of the nodes or how class labels relate to the network links. We wish to identify the best subset of nodes to label in order to learn this relationship between node attributes and network links.  We can then use this discovered relationship to accurately predict the class labels of the rest of the network nodes. 

We present a model that identifies groups of nodes with similar link patterns, which we call \textit{network roles}, using a generative blockmodel.  The model then predicts labels by learning the mapping from network roles to class labels using a maximum margin classifier. We choose a subset of nodes to label according to an iterative margin-based active learning strategy. By integrating the discovery of network roles with the classifier optimisation, the active learning process can adapt the network roles to better represent the network for node classification. We demonstrate the model by exploring a selection of real world networks, including a marine food web and a network of English words. We show that, in contrast to other network classifiers, this model achieves good classification accuracy for a range of networks with different relationships between class labels and network links. 
\end{abstract}

\maketitle

\section{Introduction}
Many naturally occurring networks can be decomposed into sets of nodes that link to the rest of the network in similar ways.  These sets of equivalent nodes  
not only share link structure but often share common attributes.  For example, assortative communities in online bidding networks correspond to the main user groups according to common interest \cite{Reichardt2007}, in blog networks blogs tend to link to others of the same political view \cite{Adamic05thepolitical}, and in biological networks proteins tend to link to others that perform similar functions \cite{ChenYuan2006}.  
In addition, there are types of disassortative link patterns where nodes that are dissimilar prefer to link to each other. For example, in a network of sexual relationships linked entities tend to be of the opposite gender \cite{Bearman2004}, species in a food web tend to eat organisms of a different species \cite{Allesina09}, and in an adjacency network of English words adjectives tend to precede nouns. 

When we encounter a new network, we may not initially know how link patterns relate to the attributes of nodes.  The network could contain relations that are assortative, disassortative or a mixture therein.  If we wish to perform learning tasks such as classification, it is important to understand the relationship between network links and node attributes.  

One way we can analyse the link patterns in networks is to identify groups of nodes that link in equivalent ways. We call these groups of nodes with similar link patterns \textit{network roles}.  By analysing the pattern of links within and between network roles we can understand the overall network structure.

Class labels, on the other hand, represent attributes or descriptions of the nodes.  Consider the aforementioned word network, in which a link indicates that words are adjacent in text and the direction of the link indicates the word order.  Let each of the words be assigned one of two possible class labels, ``noun'' or ``adjective''.  Words labelled adjective are descriptive words and words labelled noun are naming words.  Here the class label tells us something about the word, but without prior knowledge of the language, it does not tell us how it links with other words.  

Therefore, class labels tell us something about the nodes and network roles tell us about how nodes link to each other.  Our goal is to identify the relationship between the two.  

In English we can use network roles to describe the link pattern that nouns follow adjectives.  However, in the French language adjectives may come before or after a noun so while they have the same class labels, they are described by a different set of roles.  Furthermore, in French certain types of adjectives such as \textit{grand}, \textit{beau} and \textit{bon} come before the noun while others such as \textit{am\'{e}ricain}, \textit{noir}, \textit{rond} usually come after the noun.  In this case, nodes of the same class display heterogeneity as they do not all link to the network in the same way, therefore we can use multiple roles to represent the heterogeneous link patterns of this class.  

\begin{figure}%
\centering
\includegraphics[width=.9\columnwidth]{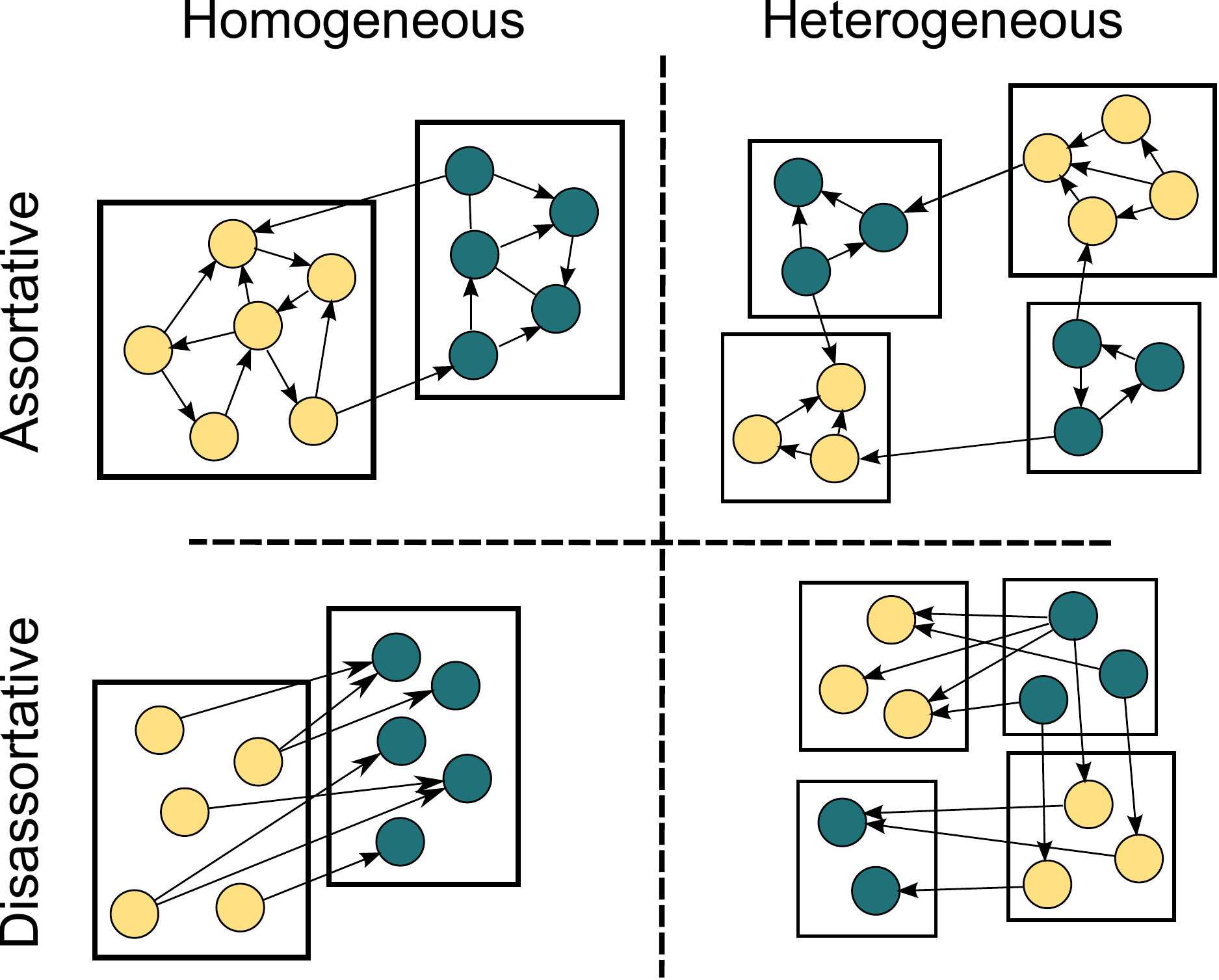}%
\caption{Four examples of classes and roles in networks. Node colour indicates class label. Nodes of that have similar link patterns are enclosed in a box; we call these network roles.  A homogeneous class is made up of a single role while a heterogeneous one contains multiple roles. }%
\label{fig:nettypes}%
\end{figure}

A graphical representation of some possible role and class combinations are shown in  Figure \ref{fig:nettypes}.  Each node has a discrete class attribute (indicated by colour). The difference between the networks is how the nodes link to the rest of the network. Nodes of the same role (i.e.\ have similar link patterns) are enclosed in a box. The top left network illustrates assortative homogeneous classes, while the network in the bottom left shows disassortative classes.  Both of these cases are homogeneous because the links of each class is described by a single role.  In contrast, the link structure of a heterogeneous class cannot be adequately represented by a single role and instead multiple roles are required to describe the link patterns of each class (right).

In this work we consider the scenario where networks links are known but the node class labels are not. Although the labels of the network nodes are not immediately available, we can query the network for more information (e.g.\ by conducting field work or lab experiments etc.).   The task is to query a subset of the labels to discover network roles, which can not only be used to understand the relationship between classes, but can also be used to predict missing class labels.  

Our approach uses only the link information and is applicable to a range of assortative and disassortative structures with no \textit{a priori} assumptions about the structure type.  We also do not assume that all nodes of a particular class link to the rest of the network in the same way.  We achieve this by using a variant of the stochastic blockmodel \cite{Holland83,nowicki2001eap} to model the probabilities of observing links between a pair of roles.  The use of a blockmodel gives us the flexibility to model a wide range of network structures, e.g.\ assortative, disassortative or a mixture of the two. It also allows for directed relations.  

The particular blockmodel variant we use allows for individual nodes to have mixed role memberships.  This means that a class contain multiple roles, and also that roles can be shared across classes.  For example, consider a predator-prey network where nodes represent species and directed links represent \textit{who-eats-whom}, we may be able to label nodes as carnivores, omnivores, herbivores and plants.  While the carnivore and herbivore classes could be represented by the distinct roles \{\textit{eats-animals}, \textit{eats-plants}\}, a node labelled as omnivore would take on a mixture of these roles.

To learn the relationship between network roles and class labels, we require some alignment of the classes and roles, i.e. so we can identify a mapping from roles to classes.  We achieve this with a blockmodel that incorporates a maximum margin classifier (e.g.\ Support Vector Machine \cite{cristianini2000introduction}).  This new model allows us to use known node labels to influence the discovery of roles to improve their alignment to the class labels.  Therefore acquiring more node labels improves the correspondence between roles and classes.  To determine which nodes to acquire to efficiently discover network roles, we employ a margin-based active learning strategy.

Once we have trained our model, we can use the network roles we have discovered to make predictions about the node labels we do not know.  This classification process is somewhat analogous to performing dimensionality reduction (e.g.\ Principal Component Analysis) on iid data before performing classification.  By representing the network with roles, we achieve similar goals to the iid case, i.e.\ reduce dimensions, remove linear correlations.  However, we also gain an additional benefit of conditional independence, since although the nodes in the network are by nature dependent on each other, once we condition on the network roles the class labels become conditionally independent of each other.

\begin{figure}
  \begin{center}
    \includegraphics[width=\columnwidth]{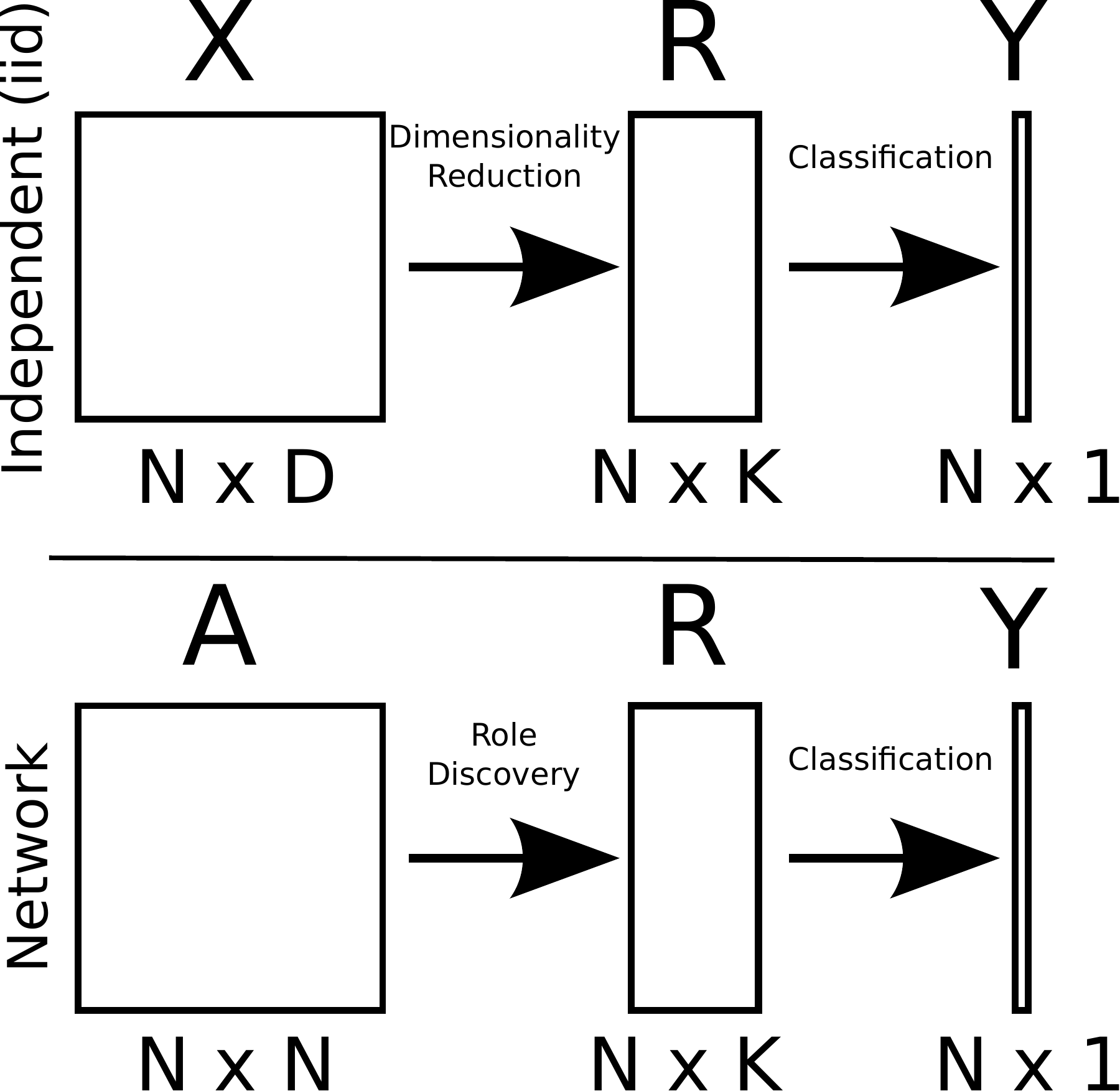}
    \caption{A comparison of an iid classification process with dimensionality reduction (top) and the network classification process of our model (bottom).  In the iid case the rows of the input feature matrix, $X$, are independent while the adjacency matrix, $A$, of the network is not.   }
    \label{fig:pipeline}
  \end{center}
\end{figure}

We test our method on a selection of real world networks and show that, in contrast to previous work, it performs well for a range of network structures and even when nodes with the same class label connect to the rest of the network in different ways.

\section{Related Work}
\label{sec:rw}
In this work we examine the problem of how to explore a network in order to discover the relationship between node labels and network links.  Related to this problem are the tasks of collective classification and active learning in networks.  These methods make predictions about node labels, but unlike our work, they do not explicitly try to identify the type of relationship between links and labels. 

The problem of classifying nodes in networks (referred to as collective classification) has received a lot of attention in recent years (e.g.\ \cite{Burfoot:2011,KongSY11,ICML2011Kuwadekar_264,LuGetoor2003,MacskassyProvost2003,sen:aimag08,Tian:2006,Zhu03combiningactive}).  These methods consider the hard problem of making a joint, or \textit{collective}, prediction about the class labels of nodes.  This is hard because for $N$ nodes with $C$ possible class labels, there are $C^N$ different ways to label the nodes collectively.  In the active learning setting, the task is to select a good subset of nodes to label in order to best predict the remaining nodes.  Of these collective classification methods, the relevant ones are those that are applicable to the univariate setting, i.e.\ they do not use attributes of the nodes as features for classification.  

Most collective classifiers make assumptions, either explicitly or implicitly, that do not necessarily hold for all networks.  The most significant one being assortativity (homophily), the is the assumption that linked nodes are more likely to be of the same class.  This assumption does not hold in all cases as not all class labels are related to the network structure in this way. The Markov assumption that node labels are conditionally independent given the labels of its neighbours (e.g.\ \cite{LuGetoor2003,MacskassyProvost2007,neville00iterative}), is also frequently used.  This assumption allows one combinatorially hard collective problem to be broken into many easier related problems that can be solved locally at the node level in conjunction with iterative algorithms to propagate these predictions around the network.  We make a similar, but less restrictive assumption that the labels of \textit{all} nodes are conditionally independent of each other, given a set of unobserved latent variables that we infer from the network structure.

\section{Role Discovery in Networks}
We present a model for discovering network roles (groups of nodes with similar link patterns) to help us understand the relationship between the class labels of nodes and the network links.  Blockmodels are a type of probabilistic generative network model and are a natural choice for this task since they can model a wide range of network structure types, e.g.\ assortative, bipartite, core-periphery, hierarchical. 

Our model is based on a mixed membership blockmodel \cite{Parkkinen_Sinkkonen_Gyenge_Kaski_2009} and therefore assumes that each node a belongs to a distribution over roles.  The nodes' distribution over roles determines the probability of a link existing between any pair of nodes. This assumption allows us to treat the links in the network as being conditionally independent of each other given the network roles.

  Since the blockmodel is a type of probabilistic generative model, we can specify it according to the assumed data generating process:
\begin{enumerate}
	\item For a given network draw a distribution over the possible $K^2$ network role interactions\footnote{Conceptually, we can think of $\mathbf{\pi}$ as a matrix with dimensions $K \times K$ such that each element $\pi_{k_1,k_2}$ represents the probability of observing a link from role $k_1$ to role $k_2$.} 
	$\pi \sim \textrm{Dirichlet}(\alpha)$
	\item For each role $k \in \{1,2,...,K\}$:
	\begin{itemize}
		\item Draw a distribution over nodes $\phi \sim$ $ \textrm{Dirichlet}(\beta) $
	\end{itemize}
	\item For each interaction $i \in \{1,2,...,I\}$:
	\begin{itemize}
		\item Draw a role-role interaction pair $z_i = (z_s,z_r)_i$, $z_i \sim \textrm{Categorical}(\pi)$
		\item Draw a sender node $s_i \sim \textrm{Categorical}(\phi_{z_s})$
		\item Draw a receiver node $r_i \sim \textrm{Categorical}(\phi_{z_r})$
	\end{itemize}
\end{enumerate}

For a given network, $G = (\mathcal{V},\mathcal{E})$, this blockmodel defines a likelihood function over the $N=|\mathcal{V}|$ nodes and $M=|\mathcal{E}|$ links or interactions.  The model assigns each link, $\{s_i,r_i\}$, in the network a latent variable $z_i$ representing a distribution over possible role pairs.  Each node's distribution over roles, $\bar{z}_v$, is then given by a normalised sum over the indicator vectors of length $K$ describing the network role of the sender ($z_{s_i}$) and receiver ($z_{r_i}$) nodes:
\begin{equation}
  \bar{z}_v = \frac{1}{n_v}\left(\sum_{i:s_i=v}{z_{s_i}} + \sum_{i:r_i=v}{z_{r_i}}\right).
\end{equation}

Exact evaluation of the blockmodel likelihood, $p(\{s,r\}|\alpha,\beta)$, is intractable since it requires a sum over all possible combinations of the latent variables, $\bf{z}$.   Using a distribution $\widetilde{q}$ and Jensen's inequality, we approximate the log likelihood with a more tractable lower bound\footnote{This is true for any choice of distribution, $\widetilde{q}$, and is an equality when $\widetilde{q}$ is equal to the posterior distribution.}:
\begin{align}
  \log  p&(\{s,r\}|\alpha,\beta) \notag\\
  & = \log \iint  \sum_{\mathbf{z}}{p(\pi,\phi,\mathbf{z}, \{s,r\}|\alpha,\beta)} \dif \pi \dif \phi \notag \\
  & = \log \iint \sum_{\mathbf{z}}{\frac{p(\pi,\phi,\mathbf{z}, \{s,r\}|\alpha,\beta)\widetilde{q}(\pi,\phi,\mathbf{z})}{\widetilde{q}(\pi,\phi,\mathbf{z})}} \dif \pi \dif \phi \notag \\
  & \geq \iint \sum_{\mathbf{z}}{\widetilde{q}(\pi,\phi,\mathbf{z}) \log \frac{p(\pi,\phi,\mathbf{z}, \{s,r\}|\alpha,\beta)}{\widetilde{q}(\pi,\phi,\mathbf{z})} \dif\pi \dif \phi} \notag \\
  & = \mathbb{E}_{\widetilde{q}}[\log p(\pi,\phi,\mathbf{z}, \{s,r\}|\alpha, \beta)] - \mathbb{E}_{\widetilde{q}}[\log \widetilde{q}(\pi,\phi,\mathbf{z})],
  \label{eq:LLlowerbound}
\end{align}
  where $\mathbb{E}_{\widetilde{q}}$ is the expectation under the $\widetilde{q}$ distribution. Since the Dirichlet and Categorical distributions are conjugate, we can solve the integration analytically to give us a tighter bound on the log likelihood \cite{TehNewWel2007}:
  \begin{align}
	\mathcal{L}(q;\alpha, \beta) &\triangleq \mathbb{E}_q[\log p(\mathbf{z}, \{s,r\}|\alpha, \beta)] - \mathbb{E}_q[\log q(\mathbf{z})] \label{eq:L} \\
	&\geq \mathbb{E}_{\widetilde{q}}[\log p(\pi,\phi,\mathbf{z}, \{s,r\}|\alpha, \beta)] - \mathbb{E}_{\widetilde{q}}[\log \widetilde{q}(\pi,\phi,\mathbf{z})]. \notag
\end{align}

\section{Supervised Role Discovery}
Our goal is to better understand the relationship between network link structure and the node class labels.  In the previous section we discussed how a blockmodel can be used to identify network roles and that network roles can be used to describe how the network links. Therefore to achieve our goal we need to find a mapping between the network roles and the class labels.  To find this mapping we use a maximum margin classifier, e.g.\ support vector machine (SVM) \cite{cristianini2000introduction}). We choose maximum margin classifiers because along with their strong generalisation guarantees \cite{Vapnik1998}, they have also demonstrated empirical success on a broad range of iid tasks. Additionally, they have been effective when used in conjunction with probabilistic models to solve problems including optical character recognition \cite{NIPS2003_AA04} and document classification \cite{Medlda}.  Effective methods for margin-based active learning in iid data have also been established \cite{Har-Peled2007, RothS06, TongK01}.   

A classifier is a function $F : \mathcal{X} \rightarrow \mathcal{Y}$ that maps a feature vector $\mathbf{x} \in \mathcal{X}$ to a label $y \in \mathcal{Y}$.  Since the blockmodel assigns each node, $v$, a vector, $\bar{z}_v$, representing its distribution over roles, we can use $\bar{z}_v$ as a the input feature vector. The classifier can then learn the relationship between roles and class labels.  In this work we restrict F to be a function of the form:

\begin{equation}
  F(\bar{z}_v,\eta) = \arg \max_{y \in \mathcal{Y}} \; \eta_y^T\bar{z}_v, 
  \label{eq:F}
\end{equation}
corresponding to a linear classifier in which $\eta$ is a matrix of coefficients with dimensions $K \times C$  and $C=|\mathcal{Y}|$ is the number of different class labels.  A classifier with good generalisation properties is one that maximises the margin, i.e.\ the distance between the training points and the separating linear decision boundary \cite{Vapnik1998}.  In the case when $C=2$ the support vector machine provides an effective method for learning a binary maximum margin classifier.  Solving the multi-class ($C>2$) case corresponds to the following constrained optimisation \cite{Crammer:2002}:

\begin{gather}
  \min_{\eta, \xi} \quad \frac{1}{2}||\eta||^2 + \frac{D}{N}\sum_{v=1}^N{\xi_v} \notag \\
  \forall v, y \quad \textrm{s.t.} : \begin{cases} \eta_{y_v}^T \bar{z}_v - \eta_y^T\bar{z}_v &\geq 1-\delta_{y,y_v} - \xi_v \\ \xi_v &\geq 0, \end{cases}
\label{eq:multisvm}
\end{gather}
where $D$ is a positive regularisation constant, $\delta_{p,q}$ is the Kronecker delta and $1-\delta_{y,y_v}$ represents our loss function, which equals $0$ for the correct prediction and $1$ for any other prediction.  The slack variables, $\{\xi_v\}$ are used for classes that cannot be separated by a linear classifier by allowing some misclassification in the training data.  Since we minimize over the $\xi_v$'s, when the classes are linearly separable then $\xi_v=0 \; \forall v$.  The regularisation parameter, $D$, controls the scale of the training misclassification penalty and can be set by using cross-validation.  
Here, the multiclass margin, by which the true class $y_v$ is favoured over another class $y$, is given by:
\begin{equation}
  \eta_{y_v}^T \bar{z}_v - \eta_y^T\bar{z}. 
\label{eq:expmargin}
\end{equation} 

The constraints in Eq.\eqref{eq:multisvm} ensure that all training instances lie on the correct side of the decision boundary, if possible, and a $\xi_v>0$ indicates that training instance $v$ is misclassified. Solving Eq.\eqref{eq:multisvm}\footnote{In practice this involves solving an equivalent dual formulation, which we omit for brevity.} requires the introduction of Lagrange multipliers, $\widehat{\mu}$, and solving the Lagrangian:

\begin{align}
  L(\eta,\xi,\widehat{\mu}) = & \frac{1}{2}||\eta||^2 + \frac{D}{N}\sum_{v=1}^N{\xi_v} \notag \\
   & + \sum_{v=1}^N{\sum_{y \in \mathcal{Y}}{ \widehat{\mu}_v^y[\eta_{y_v}^T \bar{z}_v - \eta_y^T\bar{z}_v + 1-\delta_{y,y_v} - \xi_v ]}} \notag \\
 \forall v, y \quad & \textrm{s.t.} : \widehat{\mu}_v^y \geq 0.
\label{eq:lagrangian}
\end{align}

Within this framework the simplest approach would then be: first, infer the node roles using the blockmodel, and second, train the SVM classifier using the known class labels and the inferred roles.  However, the likelihood function of blockmodels often contains a large number of local optima pertaining to good but distinct solutions \cite{good:performance}.  As a result, optimising Eq.\eqref{eq:L} might not result in the discovery of roles that relate to the class labels\footnote{In the worst case the inferred roles could be orthogonal to the class labels.}.  Instead, we introduce the maximum-margin blockmodel (MaxBM), that treats the training of the classifier and the inference of the roles as a joint optimisation problem given by:
\begin{multline}
  \min_{q,\eta, \xi} \quad - \mathcal{L}(q;\alpha,\beta) + \frac{1}{2}||\eta||^2 + \frac{D}{N}\sum_{v=1}^N{\xi_v}  \\
  \forall v, y \quad \textrm{s.t.} : \begin{cases} \mathbb{E}_q[\eta_{y_v}^T \bar{z}_v - \eta_y^T\bar{z}_v] &\geq 1-\delta_{y,y_v} - \xi_v \\ \xi_v &\geq 0, \end{cases}
\label{eq:opt}
\end{multline}
 
The first part of the optimisation corresponds to the negative lower bound on the likelihood of the blockmodel (Eq.\eqref{eq:L}), while the second part corresponds to the margin based classifier (Eq.\eqref{eq:multisvm}).  Formulating the problem like this means that the model will avoid locally optimal solutions (i.e.\ network roles) if they do not help with the classification task.  

As with other blockmodels, the MaxBM treats the presence or absence of each link as being conditionally independent given the latent role assignment.  Additionally, the model assumes that given the role assignment, the class labels are conditionally independent of each other and the network structure.

\section{Inference}
\label{sec:inference}
We fit the MaxBM model to the observed network using an expectation-maximisation (EM) style procedure (see Algorithm \ref{alg:maxbm-inf}).  In the expectation step, we infer the latent variables (i.e.\ the network roles $\mathbf{z}$), and in the maximisation step, we learn the model parameters (i.e.\ the classification coefficients $\eta$).  Once the algorithm has converged, we can use the inferred model to make predictions about the unknown node labels. 


\begin{algorithm}[t]
 \caption{Inference algorithm for Maximum-Margin Blockmodel}
 \label{alg:maxbm-inf}
 \begin{algorithmic}[H]
 \STATE Initialise $\bm{\lambda, \eta}$ randomly.
 \STATE Initialise $\bm{\widehat{\mu}} = 0$.
 \WHILE{$\textrm{relative improvement in } \mathcal{L}>10^{-6}$}
 \FOR{ $i = 1  \textrm{ to } |\mathcal{E}|$}
   \STATE update $\lambda_i$ using Eq.\eqref{eq:updatelambda}.
 \ENDFOR
 \STATE update $\bm{\widehat{\mu},\eta}$  by optimising Eq.\eqref{eq:opt}.
 \ENDWHILE
 \STATE Predict unknown labels $\widehat{y}_v$ using Eq.\eqref{eq:predNode}
\end{algorithmic}
\end{algorithm}

\subsection{Inferring the network roles}
We infer the network roles, $\mathbf{z}$, using variational Bayesian (VB) inference.  VB inference \cite{attias00avariational} has the advantage over sampling-based methods due to convergence that is faster and easier to diagnose.   In VB, a variational posterior distribution is used to lower bound the log likelihood (see Eq.\eqref{eq:LLlowerbound}).  This variational distribution is restricted to a set of tractable distributions to approximate the true posterior distribution.  Most frequently this restriction is that the distribution is fully factorised, known as a mean-field approximation.

As we have integrated out the parameters $\pi$ and $\phi$, our variational posterior distribution, $q$, is over the latent roles, $\mathbf{z}$, only.  This distribution takes the form:
\begin{equation}
	q(z) = \prod_i{q(z_i|\lambda_i)},
\label{eq:collpost}
\end{equation}   
where each $\lambda_i$ is a $|K \times K|$-dimensional variational parameter for a categorical distribution over pairs of roles for link $i$.

We optimise over $q(\mathbf{z})$ by optimising $q(\{z_i\})$ for each edge in the network until convergence.  Since our model is composed of probability distributions from the conjugate exponential family the updates take a particular general form \cite{Sung:2008:LVB}:
\begin{equation}
  q(z_i) \propto \exp (\mathbb{E}_{\sim q_{z_i}}[\log p(Y,Z)])
\label{eq:lsvbup}
\end{equation}
where $Y$ represents the observed data and $\mathbb{E}_{\sim q_{z_i}}$ is the expectation under all $q_{z_i'}$ for all $i' \neq i$.  Exact calculation of the expectation in \eqref{eq:lsvbup} is computationally expensive and therefore we use a first order Taylor expansion to approximate the update equations \cite{AsuWelSmy2009a,Sung:2008:LVB}.  The network roles are therefore updated according to:
\begin{align}
	\lambda_{i,k_1,k_2} \propto   \left( d_{k_1,k_2}^{\neg i} \right. & \left.\vphantom{d_{k_1,k_2}^{\neg i}} + \alpha_{k_1,k_2} \right) \frac{(n^{\neg i}_{s_i,k_1} + \beta)(n^{\neg i}_{r_i,k_2} + \beta)}{(n^{\neg i}_{\cdot, k_1} + N\beta)(n^{\neg i}_{\cdot,k_2} + N\beta + \delta_{k_1,k_2})}   \notag\\
	  \times \exp \left(  \vphantom{\frac{1}{n_{s_i}}\sum_y} \right. & \frac{1}{n_{s_i}} \sum_y{\mu_{s_i}^y \mathbb{E}[\eta_{y_{s_i},k_1} - \eta_{y,k_1}]}  \notag\\
  & + \left. \frac{1}{n_{r_i}} \sum_y{\mu_{r_i}^y \mathbb{E}[\eta_{y_{r_i},k_2} - \eta_{y,k_2}]} \right)
\label{eq:updatelambda}
\end{align}
\normalsize
where $d_{k_1,k_2}$ is the count of links from role $k_1$ to role $k_2$, and $n_{v,k}$ is the number of times node $v$ interacts as role $k$.  The first line of \eqref{eq:updatelambda} relates to the unsupervised part of the update. This part of the update is used to fit a blockmodel without the use of class labels  and can be used as an alternative to the Gibbs sampling method used in \cite{Parkkinen_Sinkkonen_Gyenge_Kaski_2009}.  The last two terms are due to the max-margin formulation of Eq.\eqref{eq:opt} and are non-zero for the nodes that lie on the decision boundary, i.e.\ the support vectors.   These create a bias in the model that encourages it to discover network roles that make more accurate predictions on these difficult examples.  

\subsection{Learning the classification coefficients}
As a consequence of the conditional independence of the class labels given the network roles, it is possible to use standard SVM optimisation algorithms\footnote{We use a modified version of $\rm{SVM_{multiclass}}$: \url{http://svmlight.joachims.org/svm_multiclass.html}}, such as \cite{Tsochantaridis:2005}, to obtain the optimal $\mu$ and $\eta$.  However, different to the standard SVM case the classifier inputs (network roles) are not observed fixed values, but are instead latent variables represented by our posterior distribution.  We represent our current approximation to the posterior distribution with a series of expectations, specifically:
\begin{align}
	&\mathbb{E}_q[z_i] = \lambda_i, \\
	&\mathbb{E}_q[\eta_y^T\bar{z}_v] = \frac{\eta_y}{n_v}^T\left[\sum_{i:s_i=v}{\lambda_{s_i}} + \sum_{i:r_i=v}{\lambda_{r_i}}\right],
\end{align}
where $n_{v}$ is the degree of node $v$ and $\lambda_{v}$ is a $K$-length vector representing the marginal probability of sender or receiver positions, i.e.:
\begin{align}
 \lambda_{s_i} = \left[\sum_{k_2}{\lambda_{i,1,k_2}}, \sum_{k_2}{\lambda_{i,2,k_2}}, \cdots, \sum_{k_2}{\lambda_{i,K,k_2}}\right]^T, \notag\\
 \lambda_{r_i} = \left[\sum_{k_1}{\lambda_{i,k_1,1}}, \sum_{k_1}{\lambda_{i,k_1,2}}, \cdots, \sum_{k_1}{\lambda_{i,k_1,K}}\right]^T. 
\end{align}

\subsection{Predicting class labels}
Once the inference algorithm has converged and we have an estimate of the network roles and classifier coefficients, we can then perform classification on the unlabelled nodes in the network according to:
\begin{equation}
  \widehat{y}_v = \arg \max_{y \in \mathcal{Y}} \; \eta_y^T\bar{z}_v. 
  \label{eq:predNode}
\end{equation}

\begin{figure*}
  \begin{center}
    \includegraphics[width=.4\textwidth]{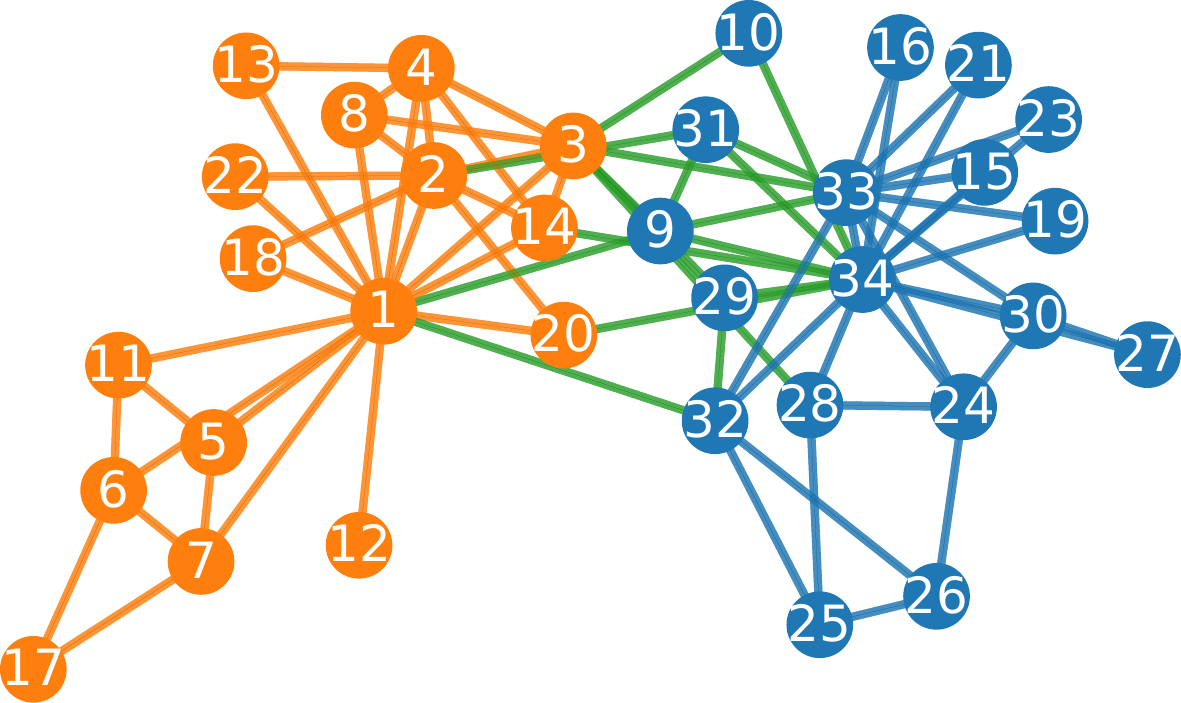}
    \includegraphics[width=.33\textwidth]{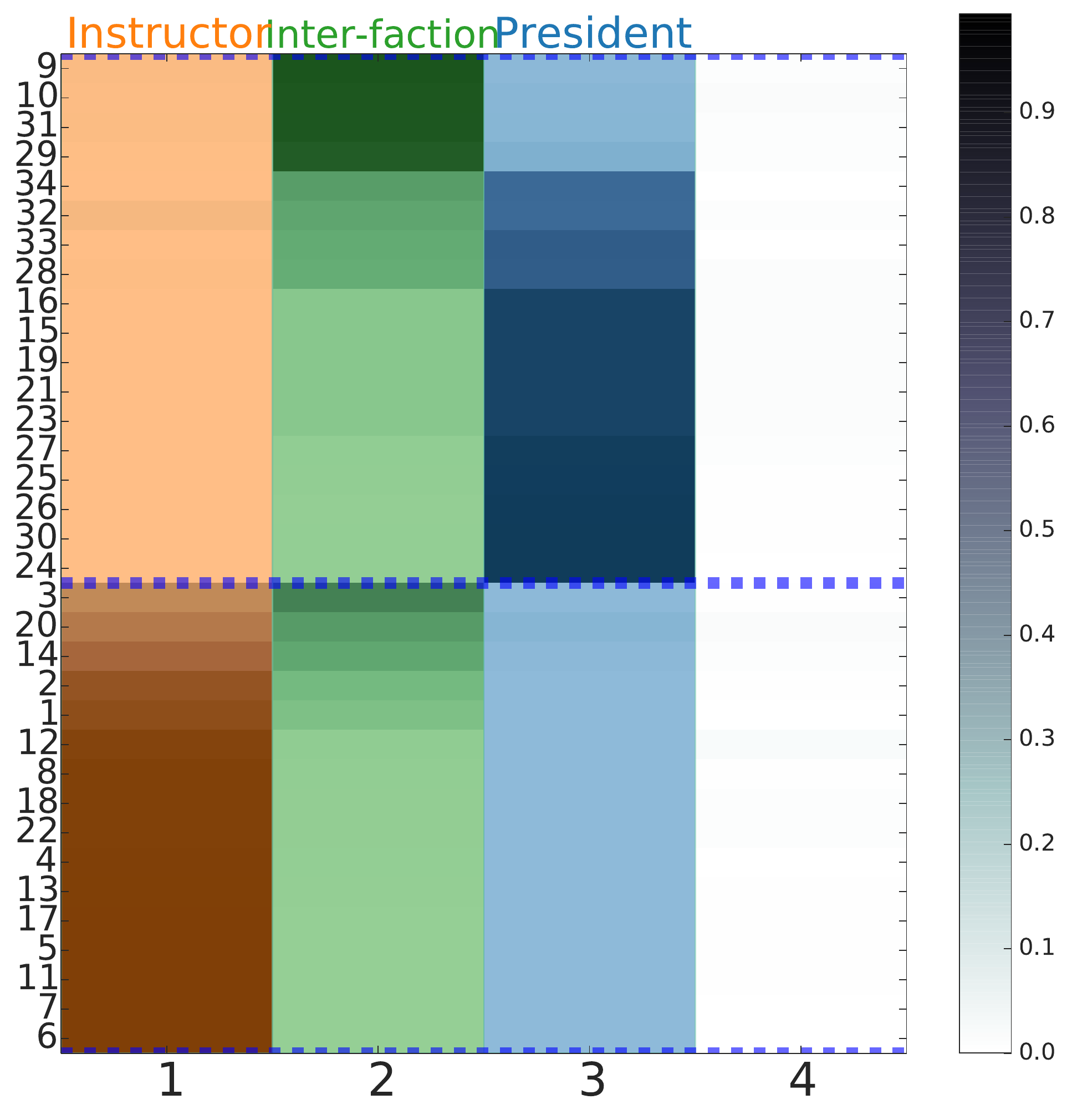}
    \includegraphics[width=.25\textwidth]{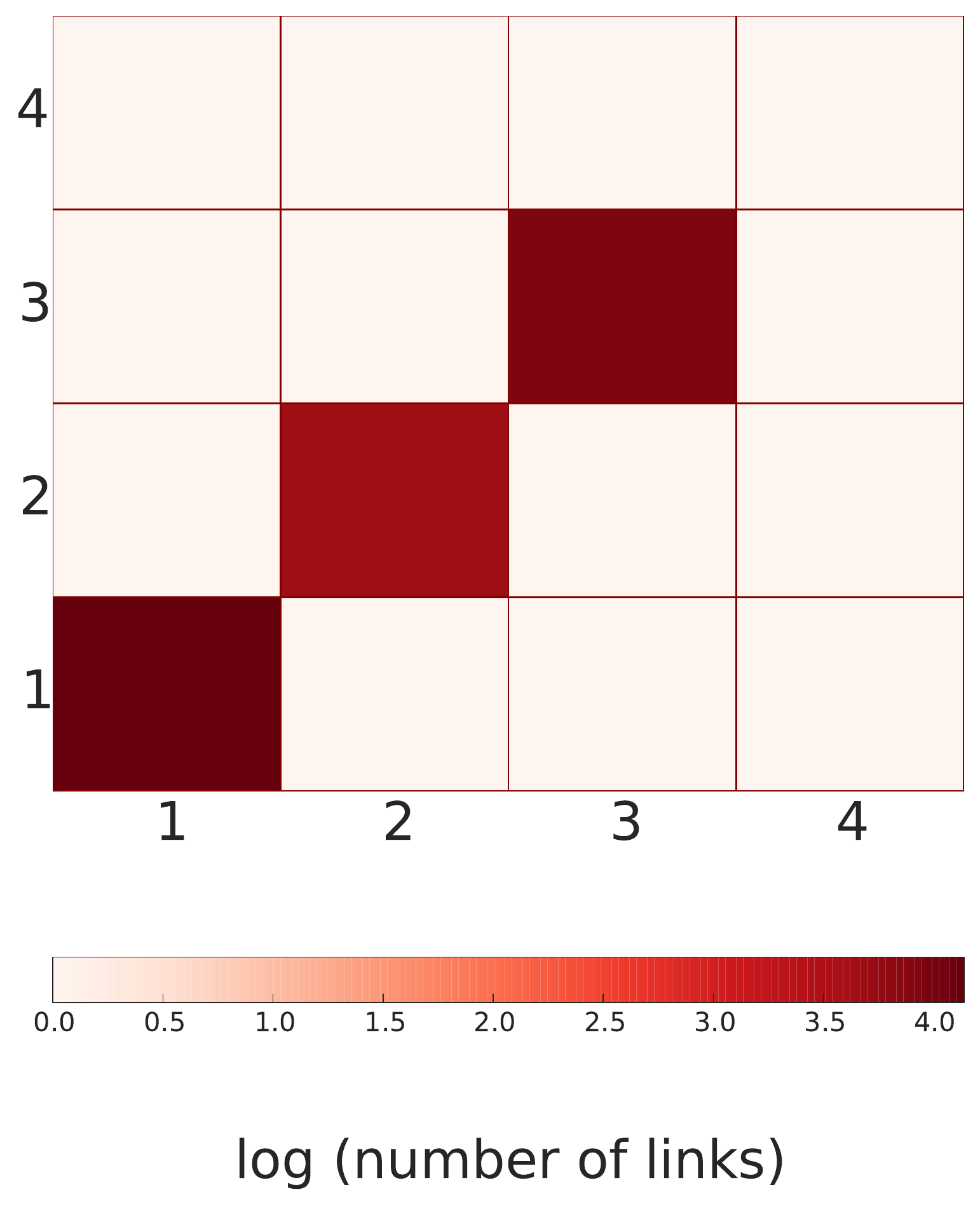}
    \caption{\textit{Left:} The karate club network.  The node colour reflects the class label, i.e.\ which faction the node belongs to, while the link colour reflects the assignment of roles.  \textit{Middle:} A visual representation of the distribution of inferred roles (columns) for each node (rows) in the network.  \textit{Right:} The role interaction matrix.  The diagonal blocks indicate that the network is assortative.}
    \label{fig:karateblocks}
  \end{center}
\end{figure*}

\section{Active Learning}
\label{sec:actlearn}
In many practical scenarios, the labels of the network nodes may not be immediately available at training time.  Acquisition of these labels involves some cost; either the cost of consulting an expert, conducting an investigation or carrying out laboratory experiments.  In order to minimise the cost incurred, we wish to select the most informative examples.  This is done using active learning.  
Active learning is a form of supervised learning that involves interactively selecting the nodes to label as part of the learning process. In this setting, network nodes start off unlabelled and the active learning algorithm aims to choose the best subset of nodes to label in order to improve the classification of the remaining nodes.  This is undertaken in a greedy manner.  At each stage of the algorithm  a single unlabelled node is selected to explore and is added to the training set.  This selection occurs based on the output of the classifier and an active learning criterion.  

As our method incorporates a maximum margin classifier it is possible to take advantage of active learning criteria developed for support vector machines.  This involves querying a sample in relation to the decision boundary.  
We employ a simple strategy that chooses nodes that have the smallest margin \cite{TongK01} as intuitively these represent the examples where the classifier is most uncertain.   We use the multiclass margin \cite{RothS06} as the query function:
\begin{equation}
  \mathcal{Q}_{\rm{multiclass}} = \arg \min_{v \in \{1,...,N\}} \eta_{\widehat{y}_v}^T \bar{z}_v - \eta_{\widetilde{y}_v}^T\bar{z}_v.
\label{eq:qmulti}
\end{equation} 
where $\widehat{y}_v$ represents the predicted class for node $v$, as given in \eqref{eq:predNode}, and $\widetilde{y}_v = \arg \max_{y \neq \widehat{y}} \eta_{y}^T\bar{z}_v$ is the second most likely class label.  The query function \eqref{eq:qmulti} therefore selects the node that has the smallest margin between the top two competing class labels.

\section{Experimental Results}
Four real world networks were used to test our Max-BM approach to learning the relationship between network links and node classes:\footnote{A Python implementation of the Max-BM model along with the datasets used are available at \url{http://gdriv.es/letopeel/code.html}} a social network, a word network, a marine food web,  and a citation network.  

The first dataset is Zachary's Karate Club \cite{Zachary_JAR1977}, a social network of friendships between 34 members of a karate club.  The club split into two factions, one led by the instructor and the other by the club president.  The class labels correspond to the  faction each node belonged to after the split.  

The second network is comprised of 112 frequently occurring nouns and adjectives in the novel \textit{David Copperfield} by Charles Dickens \cite{Newman:words}. It is a directed network where the links indicate adjacent words and the order they appear in the text.  

The third network is a food web of consumer-resource interactions between 488 species in the Weddell Sea \cite{foodweb}.  The directed edges link each predator to its prey.  We perform two classification tasks on this dataset, according to the attributes feeding type and habitat; these attributes partition the network in different ways.  Feeding type takes one of $C=6$ classes: primary producer, carnivorous, carnivorous/necrovorous, detrivorous, herbivorous/detrivorous and omnivorous.  Habitat has $C=5$ classes, namely benthic, benthopelagic, demersal, land-based, and pelagic.  

Finally, the fourth network is the much studied Cora citation network \cite{sen:aimag08} containing 2708 machine learning papers with directed links indicating citations, and $C=7$ class labels indicating one of 7 subject areas. 

For each network the algorithm was initialised by acquiring the label of one randomly chosen node from each class. It would be possible to learn the number of classes as the network is explored by adding a new class whenever a new label was encountered.  It would also be possible to optimise the number of network roles using cross-validation at each stage.  In this work, however, the number of roles was fixed, since we found in practise that as long as we had a sufficient number of roles, the exact setting of this parameter did not have a substantial effect on the results.  The reason for this is that the model only uses the roles it needs to represent the link patterns that correspond to classes (see Sec.\ \ref{sec:karate}).  We therefore set the number of network roles to twice the number of classes, i.e.\ $K = 2C$. 

Each stage of the learning process consisted of carrying out the inference procedure in Section \ref{sec:inference} to convergence, followed by the selection of a new node to be labelled.  At each stage the margin cost parameter $D$ was optimised using cross-validation.  The reported results are averaged over 50 random initialisations.

In the following, we examine the network roles discovered and the performance of the algorithm in discovering these roles.

\subsection{Role Discovery}
\label{sec:roledisc}
When confronted with a new network dataset it is important to understand the patterns of complex interactions between the nodes. This is useful to further our  understanding of how a system works, and also to better understand the data so that we can build better algorithms for prediction.  In this section we examine the network roles found using the MaxBM model and how they can be used to understand the structure of the network in relation to the attributes of nodes. 

First, we consider the karate club network.  This is a well-studied network, largely due to its simple structure and  small size.  Figure \ref{fig:karateblocks} (left) shows a visualisation of the network.  The node colours indicate the factions (which we use as class labels) to which the nodes belong to after the split.  Node 1 and node 34 represent the Instructor and club President respectively.  In Figure \ref{fig:karateblocks} (right) we see a visualisation of the logarithm of the number of links that occurs between the roles.  All the links occur on the diagonal so this tells us that the network links are assortative.  

Figure \ref{fig:karateblocks} (middle)  shows the node memberships to roles.  In this visualisation, the rows correspond to nodes in the network and the columns correspond to the $K=4$ roles.  The rows in the top half are the nodes in the President's faction, while the rows in the bottom belong to the Instructor's faction.  Notice that the fourth column does not contain any membership.  Although the model had 4 roles available, it determined that three were sufficient to represent the network structure.  
This visualisation shows the relationship between roles and classes.  We can also interpret the roles in terms of the link patterns they represent.  Role 1 and Role 3 represent interaction within the Instructor and President factions respectively, while Role 2 represent interactions across factions.  

\begin{figure}
\begin{minipage}{\columnwidth}
  \begin{center}
    \includegraphics[width=.55\columnwidth]{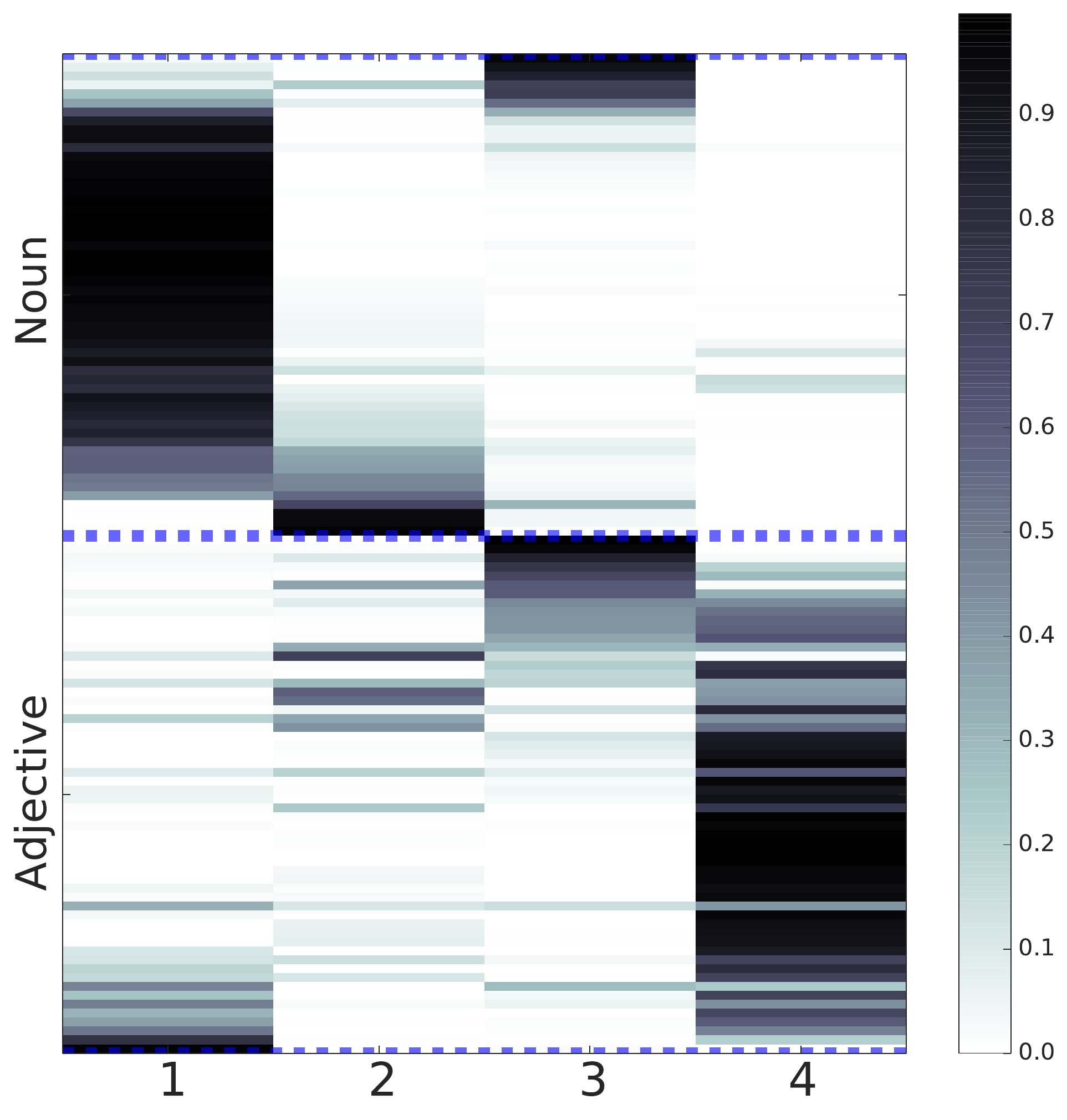}
    \includegraphics[width=.43\columnwidth]{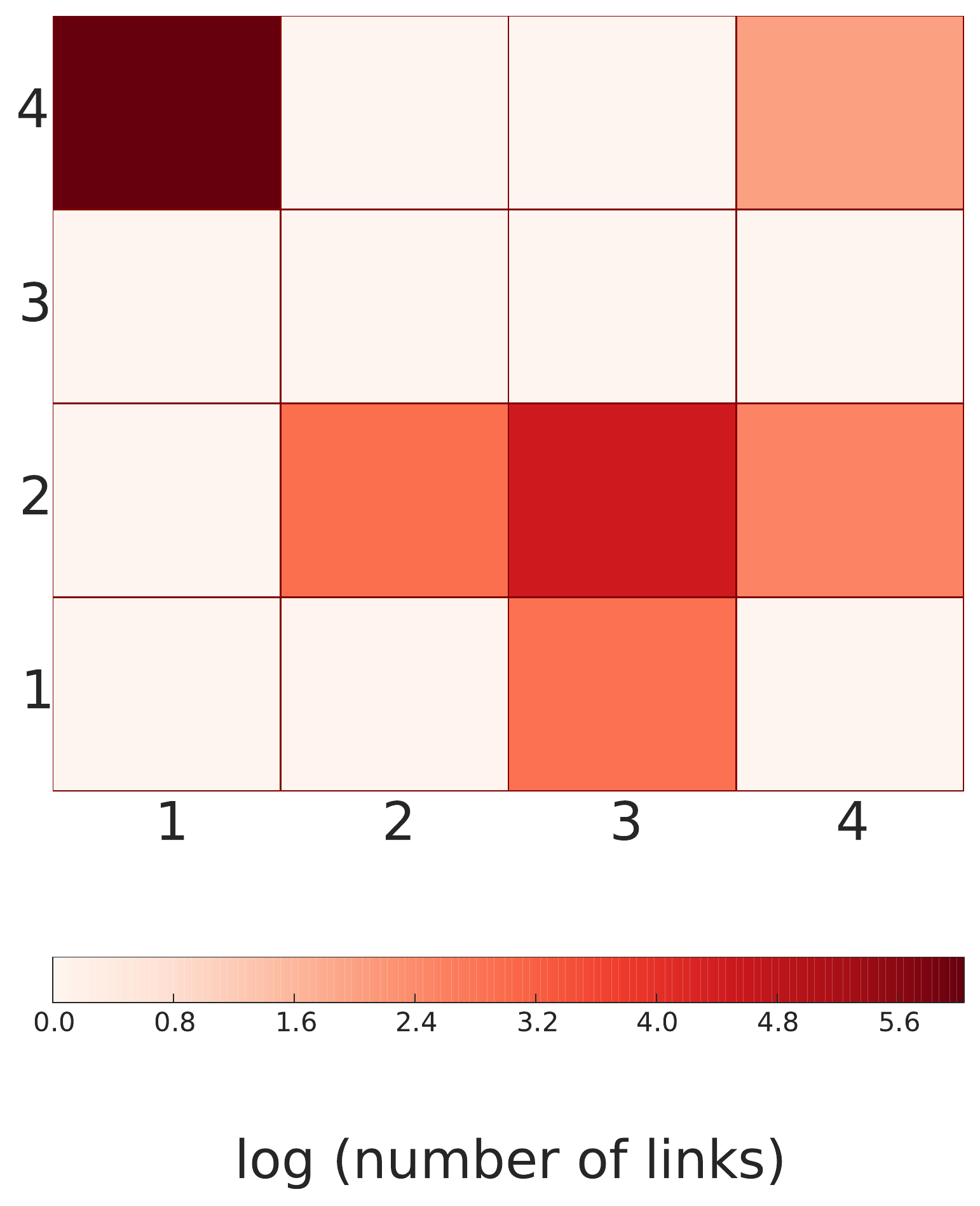}
    \caption{\textit{Left:} The distribution of inferred roles (columns) for the nodes (rows) in the word network.  \textit{Right:} The role interaction matrix.}
    \label{fig:wordblocks}
  \end{center}
\end{minipage}
\begin{minipage}{\columnwidth}
  \begin{center}
    \includegraphics[width=.55\columnwidth]{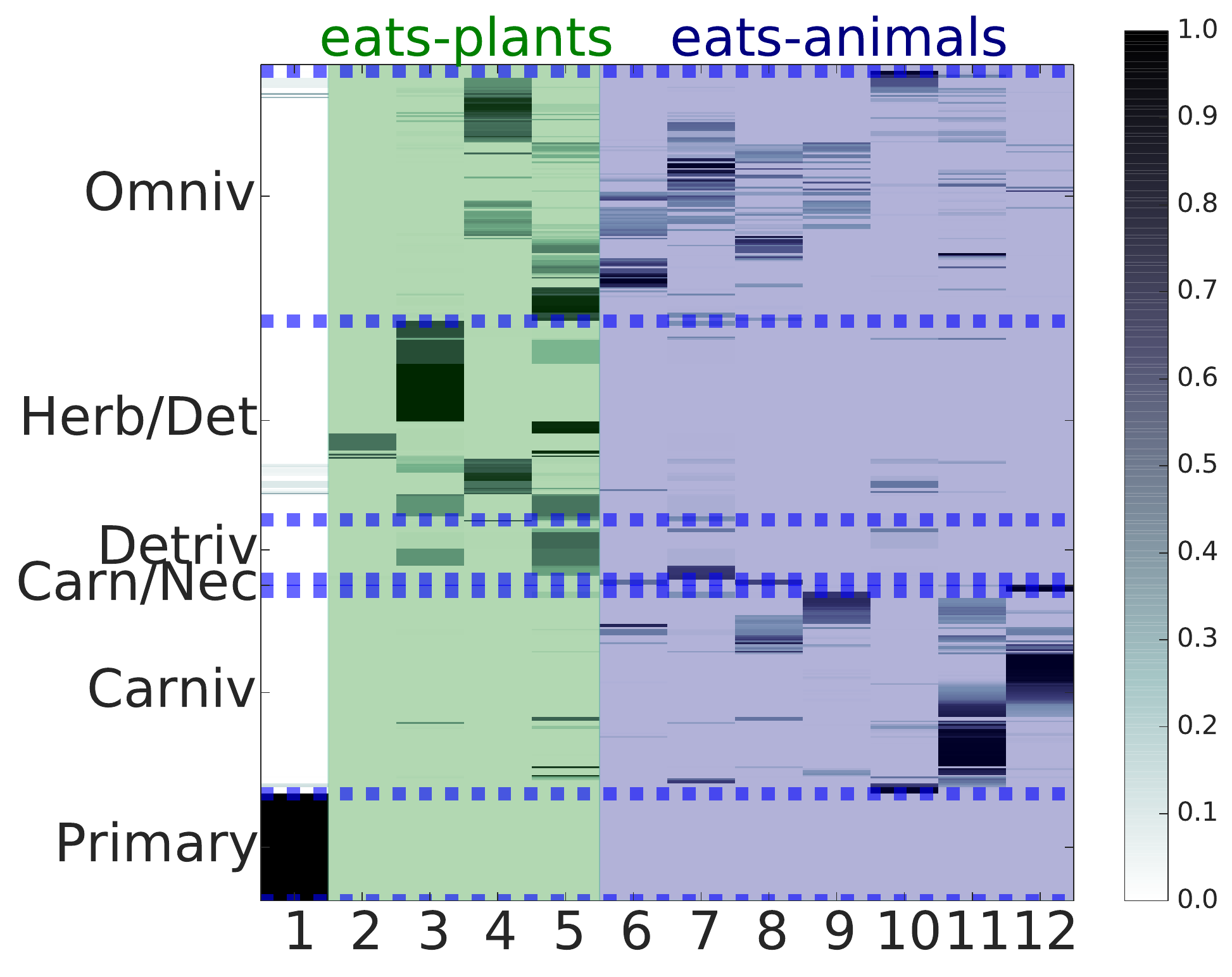}
    \includegraphics[width=.43\columnwidth]{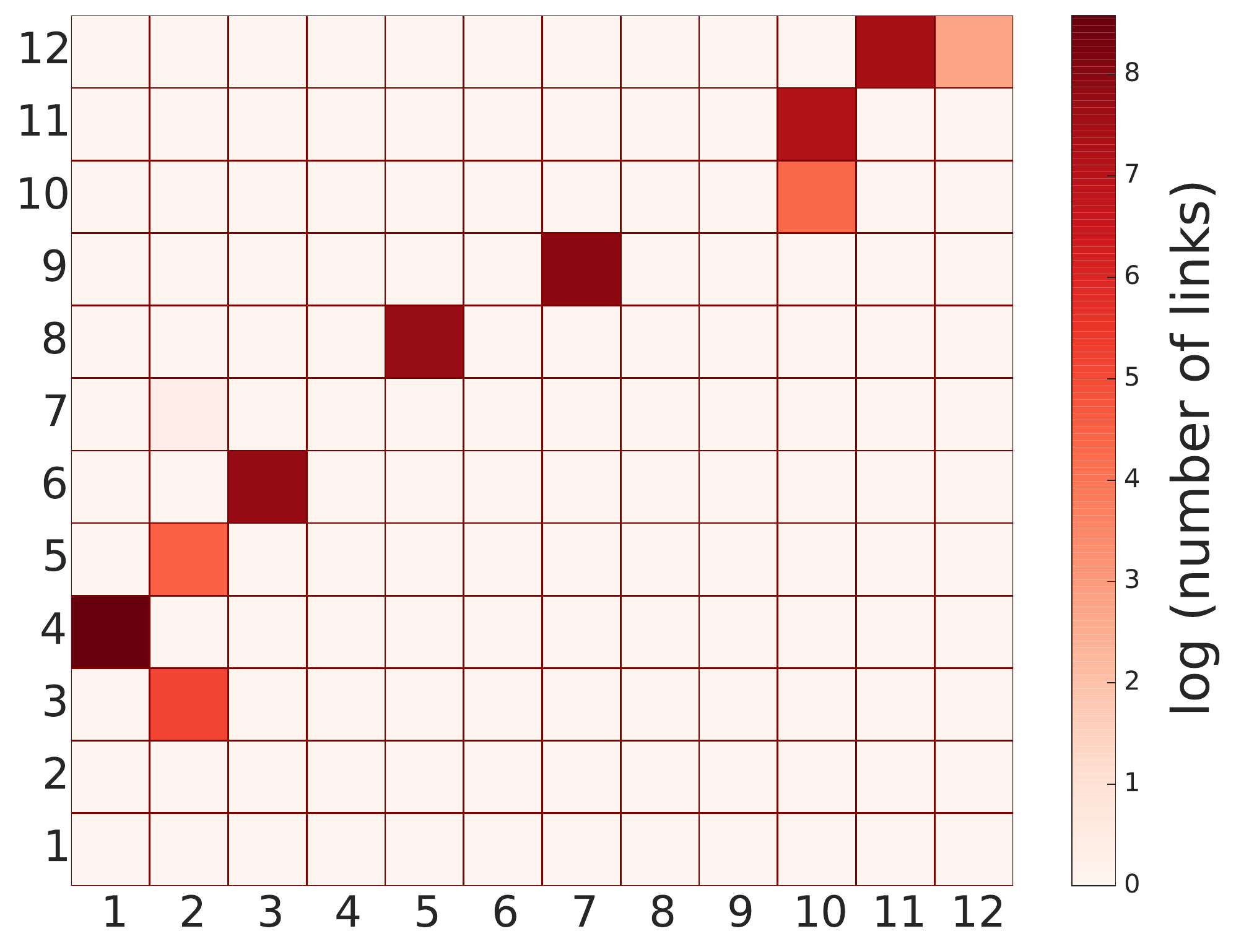}
    \caption{\textit{Left:} The distribution of inferred roles (columns) for the nodes (rows) in the food web network. \textit{Right:} The role interaction matrix.}
    \label{fig:sfwblocks}
  \end{center}
  \end{minipage}
\end{figure}

The second network is the word network of adjectives and nouns.  Figure \ref{fig:wordblocks} shows the inferred role memberships (left) and role interactions (right).  This is a network that has been previously described as being approximately bipartite \cite{Newman:words,Moore:2011} due to the tendency for adjectives to precede nouns in the English language. We observe this relationship in Roles 1 and 4, which account for the majority of the nouns and adjectives respectively.  Roles 2 and 3 are therefore used by the model to account for the words that do not follow this simple rule.  

The third network is the food web of consumer-resource interactions (i.e.\ who-eats-whom) and is shown in Figure \ref{fig:sfwblocks}. As there are two classification tasks we choose to examine the network roles in relation to the feeding type class label.  This is because it is the easier of the two to interpret without the use of more detailed domain specific knowledge. We see in Fig.\ \ref{fig:sfwblocks}(right) that the majority of the interactions lie on the off-diagonal, indicating that the network is disassortative.  This seems reasonable as it tells us that in general, species of one type tend to eat species of a different type.  Looking closer at the role memberships (Fig.\ \ref{fig:sfwblocks}(left)), we see how the class labels relate to network roles.  We see that Role 1 is exclusively composed of primary producers.  Roles 2 to 5 (green overlay) correspond to herbivore species and Roles 6 to 12 (blue overlay) correspond to carnivore species.  We might therefore think of these roles as being \textit{eats-plant}s and \textit{eats-animals} respectively and the individual roles being another level of organisation within these roles. The omnivore species are distributed across all of these roles, since we know that omnivores eat both plants and animals.

\begin{figure*}[t]
\begin{minipage}{\textwidth}
\centering
\includegraphics[width=\textwidth]{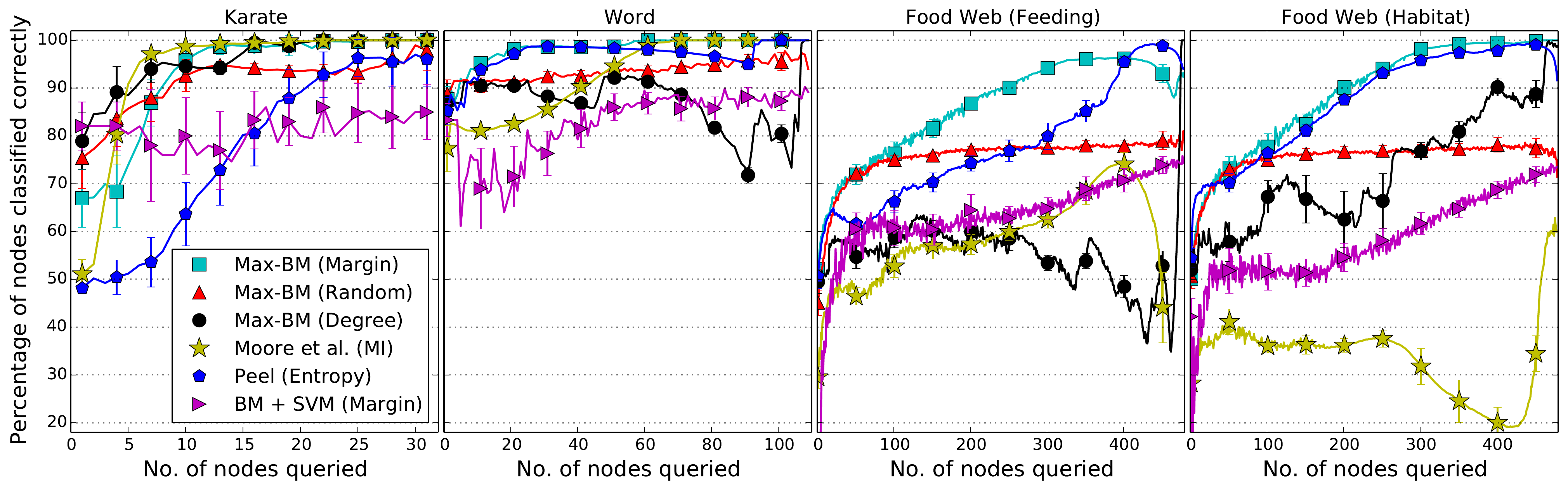}%
\caption{A comparison of the performance  of labelling nodes for the blockmodel-based approaches.}%
\label{fig:perfBM}
\end{minipage}
\begin{minipage}{\textwidth}
  \begin{center}
    \includegraphics[width=\textwidth]{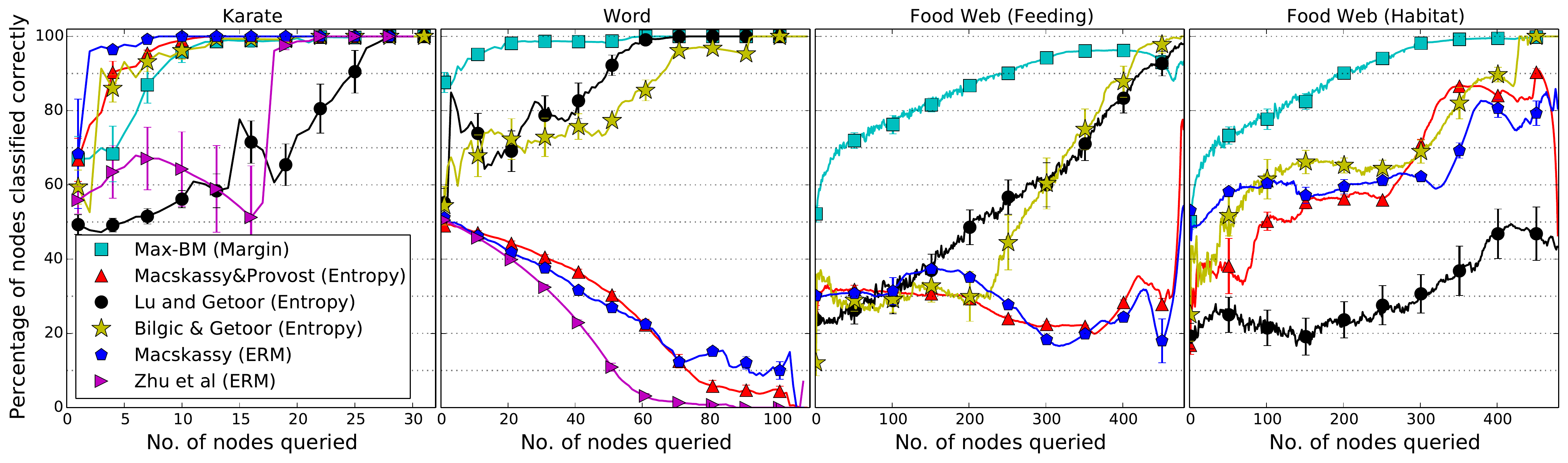}
    \caption{Performance of labelling nodes using collective classifier approaches.}
    \label{fig:perfCC}
  \end{center}
  \end{minipage}
\end{figure*}

\subsection{Classification Performance}
We showed in the previous section that the MaxBM can be used to discover the pattern between network links and node attributes.  Now we examine the performance of the model in discovering this pattern as we go through the process of acquiring labels.  
Each time we acquire the label for a new node, we quantify the performance according to the accuracy of the model predictions on the rest of the network.

We compare our algorithm (MaxBM+Margin) with two baseline strategies: \textit{random}, where nodes were queried in a random order, and \textit{degree}, where nodes where queried in order of largest degree.  We also compare against three other blockmodel approaches: a standard blockmodel using mutual information strategy(\textit{Moore et al.(MI)}) as described in \cite{Moore:2011},  the blockmodel of \cite{Peel:Fusion11} using entropy (\textit{Peel (Entropy)}), and a SVM using the roles discovered using an unsupervised blockmodel \cite{Parkkinen_Sinkkonen_Gyenge_Kaski_2009} (\textit{BM+SVM (Entropy)}).

In addition, we also compare against a selection of univariate collective classification and network based active learning methods from the literature.  
A summary of these classification and active learning selection strategies is given below\footnote{We use the netkit \cite{MacskassyProvost2007} implementations available at \url{http://netkit-srl.sourceforge.net/}}:
\begin{itemize}
  \item \textit{Macskassy \& Provost (Entropy)}: The weighted vote relational neighbour classifier \cite{MacskassyProvost2003} and selecting the node that has the highest entropy prediction.
  \item \textit{Lu \& Getoor (Entropy)}:  The network-only version of the count-link logistic regression classifier \cite{LuGetoor2003} used with the above entropy method.
  \item \textit{Bilgic \& Getoor (Entropy)}:  The naive Bayes classifier and selecting the node with the highest sum of entropy over its neighbours and itself \cite{bilgic:nips09-wkshp}.
  \item \textit{Zhu et al. (ERM)}: The Gaussian harmonic classifier using empirical risk minimisation (ERM) to select the next node \cite{Zhu03combiningactive}.
  \item \textit{Macskassy (ERM)}:  The same as above but using heuristics to select a subset of nodes to evaluate using the ERM method \cite{Macskassy2009}.
\end{itemize}

\subsubsection{Karate club network}
\label{sec:karate}
Figures \ref{fig:perfBM} and \ref{fig:perfCC}(outer left) show the performance of the different algorithms on the karate club network.  It can be seen that although our margin-based approach initially performs worse than random, after exploring about 10 nodes it has almost perfect classification accuracy.  
However, we see that a number of the other approaches out-perform our method, particularly the \textit{Macskassy \& Provost(Entropy)} and \textit{Macskassy (ERM)} approaches.  The reason for this being that these models have strong assortativity assumptions and, as we showed in the previous section, this is an assortative network.  While our approach needs to learn the structure of the network, these collective classifiers only require a good choice of nodes to make accurate predictions.

\subsubsection{Word network}

Figures \ref{fig:perfBM} and \ref{fig:perfCC}(centre left) show, at each stage, the proportion of unexplored nodes in the word network that are labelled correctly by each of the methods.  Here it can be seen that the accuracy of the \textit{MaxBM (Margin)} method quickly rises to 90\% and after querying just 20 nodes is close to 100\% accurate.  To start with, the blockmodel method, \textit{Peel (Entropy)}, performs almost as well as the margin approach, but the performance drops slightly after exploring about half of the network.  We can also see that the MaxBM method performs better than all the other approaches, since those that achieve 100\% accuracy, require the labels of at least half of the network before they do.  

In \cite{Moore:2011} they observe that, because adjectives in the English language tend to precede nouns, nodes with high out-degree can be classified as adjectives and nodes with high in-degree as nouns. Consequently they find that their mutual information approach, \textit{Moore et al. (MI)}, tends to query nodes with almost equal in- and out- degree first; these represent node labels about which the model is most uncertain.  

We performed a similar analysis using our MaxBM approach.  We found that the MaxBM quickly discovered the main roles representing the nouns that followed adjectives, i.e.\ Roles 1 and 4 in Fig.\ \ref{fig:wordblocks}; these nodes tended to be queried last.  With Roles 1 and 4 in place, the model chooses nodes that don't fit this pattern to determine how to best assign the remaining roles.  

To understand why the algorithm chose this ordering, we considered the degree of each node not only in terms of its in- and out- degree, but in terms of its in- and out- degree \textit{to each class}.  We found that the majority of the adjectives in the network not only had a higher out-degree than in-degree, but also a higher out-degree to nouns than out-degree to adjectives.  That is to say, these adjectives preceded nouns more than they preceded other adjectives.  These nodes tended to be queried last.  In contrast, adjectives that did not link to the rest of the network in this way were queried earlier since the model was more uncertain about their classification.  A similar analysis of the nouns shows that the subset of nouns explored last never follow nouns and have a relatively high in-degree.

\subsubsection{Food web}
In the food web network we have two classification tasks.  The feeding type task is shown in Figures \ref{fig:perfBM} and \ref{fig:perfCC}(centre right) and the habitat task is in Figures \ref{fig:perfBM} and \ref{fig:perfCC}(outer right).  We did not run the \textit{Zhu et al. (ERM)} method on this network due to the long running time.  For both tasks the margin strategy with the MaxBM model and the \textit{Peel (Entropy)} methods outperform the other algorithms.  

Again, we examine the query order under the \textit{MaxBM (Margin)} for both classification tasks.  For the classification of feeding type, the algorithm tends to query primary producers last and the omnivores early on.  As we saw in Fig.\ref{fig:sfwblocks}, the primary producer is a homogeneous class and is represented as a role defined as having no outgoing links (Role 1).  Omnivores, on the other hand, are the hardest to distinguish from the rest of the network as they have the greatest variation in link patterns. 

For the habitat classification task, the last half of the network nodes to be explored tend to be of the benthic and pelagic classes.  In \cite{Moore:2011} it was suggested that the diversity of the species contributed to the misclassification of more than half of the benthic species.  In contrast to previous work, the evidence suggests that our approach performs well even when the diversity is large.  The MaxBM model can capture this diversity because it allows for class heterogeneity, i.e.\ the model can use multiple roles to model the variety of link patterns within  a class.  Furthermore, the fact that the benthic nodes are queried later on suggests that not only does this model predict the benthic class labels more accurately, but it does so with greater confidence.

\subsubsection{Cora citation network}
\begin{figure}
  \begin{center}
    \includegraphics[width=.8\columnwidth]{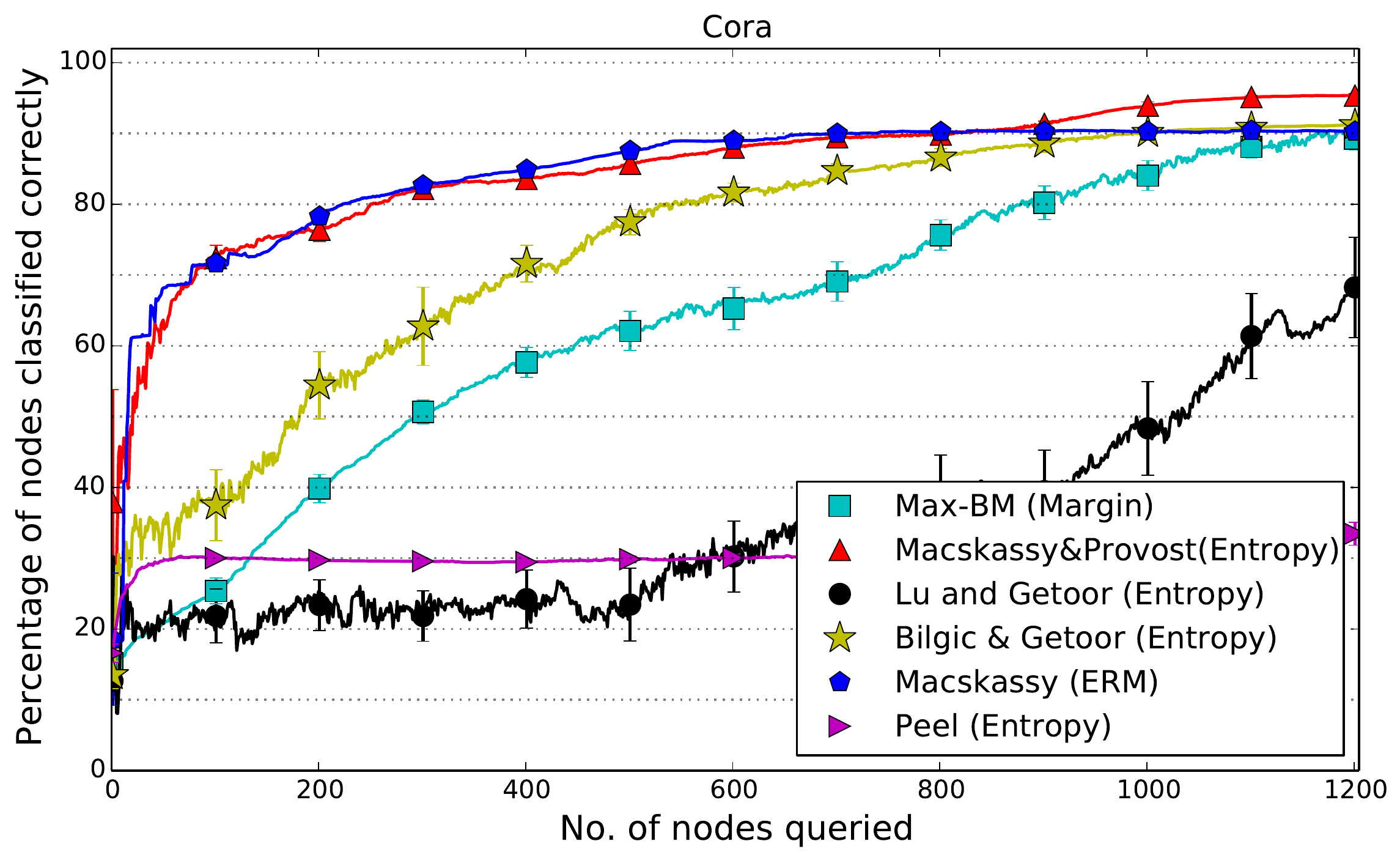}
    \caption{Performance of labelling nodes within the Cora citation network.}
    \label{fig:coraacc}
  \end{center}
\end{figure}

We do not run the \textit{Moore et al.(MI)} or the \textit{Zhu et al.(ERM)}  learning algorithms as they do not scale well to networks of this size.  Figure \ref{fig:coraacc} shows the accuracy of predicting the unlabelled nodes as the network is explored.  Unlike for the other networks, we do not explore the entire network and instead terminate the learning process once half of the nodes have been queried.  

We see that the \textit{Peel (Entropy)} method performs by far the worst.  It can be seen that the \textit{MaxBM (Margin)} method performs better, but it is the \textit{Macskassy \& Provost(Entropy)} and \textit{Macskassy (ERM)} methods that perform best. By the time half of the network has been explored all of the methods except \textit{Peel (Entropy)} and \textit{Lu \& Getoor (Entropy)} reach 90\% accuracy.  This is a similar pattern to that observed in the karate club network.  Both the karate club social network and the Cora citation network are assortative and homogeneous \cite{Peel:Fusion11}.

\vspace{3mm}
\noindent\textit{Benefit of supervised role discovery}

Finally, we make a general comment on the benefit gained by allowing the class labels to influence the role discovery process.  The \textit{BM+SVM} method is similar to our MaxBM method, except that it performs the role discovery and classifier training as  separate steps and the role discovery does not use the known class label information.  In all five of the classification tasks we see that the \textit{BM+SVM} method performed worse than the MaxBM model.  This shows that the network roles are adapting as a result of the acquisition of class labels during the active learning process and that there is measurable benefit to incorporating labels into the inference process.

\section{Discussion}
In this work we have considered the problem of identifying the best subset of nodes to label, in order to discover network roles that describe the relationship between network links and node labels.  We constructed an interpretable  and flexible model based on blockmodels and margin-based classification.  We assessed its performance at discovering roles by its ability to accurately predict the labels across the rest of the network. 

We build on previous work based on generative blockmodels and active learning to explore an unlabelled network.  In contrast to previous work, we do not assume that nodes with the same label all connect to the rest of the network in the same way.  By allowing for heterogeneity within classes we can model a wider range of network structures, while still attaining good classification accuracy on networks with simple class structures. 

We have compared our model, based on classification accuracy, to the related class of models known as collective classifiers.  We see that for simple assortative networks, some of these algorithms outperform ours.  If we know \textit{a priori} how classes in a network are distributed relative to their link structure \textit{and} we know that this relationship is assortative, then our method is not the method to use.  However, if the relationship between class labels is yet to be discovered then, as we have demonstrated, our MaxBM model provides a good approach to explore the network.  Furthermore, if the link patterns are not assortative, then our method gives better classification performance than previous methods.

\section{Acknowledgments}
We thank Chris Aicher, Mustafa Bilgic, Aaron Clauset, Abigail Jacobs, Dan Larremore, Sofus Macskassy and Suzy Moat for helpful comments and Cris Moore and Yaojia Zhu for useful discussions and providing the Weddell Sea dataset.  
The author was supported by the UK EPSRC-funded Eng. Doctorate Centre in Virtual Environments, Imaging and Visualisation (EP/G037159/1)
%

\end{document}